\documentclass{article}

\usepackage{microtype}
\usepackage{graphicx}
\usepackage{subfigure}
\usepackage{booktabs} % for professional tables
\usepackage{amsmath}
\usepackage{hyperref}

\usepackage[accepted]{icml2018}

\usepackage{algorithmic}
\usepackage{booktabs}       % professional-quality tables
\usepackage{ amssymb }
\usepackage{float}
\usepackage{bbm}

\DeclareMathOperator*{\softmax}{softmax}

\DeclareMathOperator*{\relu}{ReLU}

\begin{document}

\twocolumn[
\icmltitle{Rapid Adaptation with Conditionally Shifted Neurons}



\icmlsetsymbol{equal}{*}

\begin{icmlauthorlist}
\icmlauthor{Tsendsuren Munkhdalai}{maluuba}
\icmlauthor{Xingdi Yuan}{maluuba}
\icmlauthor{Soroush Mehri}{maluuba}
\icmlauthor{Adam Trischler}{maluuba}
\end{icmlauthorlist}

\icmlaffiliation{maluuba}{Microsoft Research, Montr\'{e}al, Qu\'{e}bec, Canada}

\icmlcorrespondingauthor{Tsendsuren Munkhdalai}{tsendsuren.munkhdalai@microsoft.com}

\icmlkeywords{Machine Learning, Meta Learning}

\vskip 0.3in
]



\printAffiliationsAndNotice{}  % leave blank if no need to mention equal contribution

\begin{abstract}
We describe a mechanism by which artificial neural networks can learn rapid adaptation -- the ability to adapt on the fly, with little data, to new tasks -- that we call \textit{conditionally shifted neurons}.
We apply this mechanism in the framework of metalearning, where the aim is to replicate some of the flexibility of human learning in machines.
Conditionally shifted neurons modify their activation values with task-specific shifts retrieved from a memory module, which is populated rapidly based on limited task experience.
On metalearning benchmarks from the vision and language domains,
models augmented with conditionally shifted neurons achieve state-of-the-art results.
\end{abstract}

\section{Introduction}
The ability to adapt our behavior rapidly in response to external or internal feedback is a primary ingredient of human intelligence. This cognitive flexibility is commonly ascribed to prefrontal cortex (PFC) and working memory in the brain. Neuroscientific evidence suggests that these areas use incoming information to support task-specific temporal adaptation and planning \cite{stokes2013dynamic,siegel2015cortical,miller2015working,brincat2016prefrontal}. This occurs on the fly, within only a few hundred milliseconds, and supports a wide variety of task-specific behaviors \cite{monsell2003task,sakai2008task}.

On the other hand, most existing machine learning systems are designed for a single task. They are trained through one optimization phase after which learning ceases. Systems built in such a \textit{train-and-then-test} manner do not scale to complex, realistic environments: they require gluts of single-task data and are prone to issues related to distributional shifts, such as catastrophic forgetting \cite{srivastava2013compete,goodfellow2013empirical,kirkpatrick2016overcoming} and adversarial data points \cite{szegedy2013intriguing}.

There is growing interest and progress in building flexible, adaptive models, particularly within the framework of \emph{metalearning} (learning to learn)~\citep{mitchell1993explanation,andrychowicz2016learning,vinyals2016matching,pmlr-v70-bachman17a}.
The goal of metalearning algorithms is the ability to learn new tasks efficiently, given little training data for each individual task.
Metalearning models learn this ability (to learn) by training on a distribution of related tasks.

In this work we develop a neural mechanism for metalearning via rapid adaptation that we call \emph{conditionally shifted neurons}.
Conditionally shifted neurons (CSNs), like standard artificial neurons, produce activation values based on input from connected neurons modulated by the connection weights. Additionally, they have the capacity to shift their activation values on the fly based on auxiliary conditioning information.
These conditional shifts adapt model behavior to the task at hand.%\footnote{We plan to release our code upon publication.}

A model with CSNs operates in two phases: a description phase and a prediction phase.
Assume, for each task $\tau \sim p(\tau)$, that we have access to a description $D_\tau$. In the simplest case, this is a set of example datapoints and their corresponding labels: $D_\tau = \{ (x'_i,y'_i) \}^n_{i=1}$.\footnote{More abstractly, a task description could be given by a set of instructions or demonstrations of expert behavior.}${}^,$\footnote{In a $C$-way, $k$-shot classification task, $n = k \times C$.}
In the description phase, the model processes $D_\tau$ and extracts conditioning information as a function of its performance on $D_\tau$.
Based on this information, it generates activation shifts to adapt itself to the task and stores them in a key-value memory.
In the prediction phase, the model acts on unseen datapoints $x_j \in \tau$, from the same task, to predict their labels $y_j$.
To improve these predictions, the model retrieves shifts from memory and applies them to the activations of individual neurons.
During training, the model learns the meta procedure of \emph{how} to extract conditioning information in the description phase and generate useful conditional shifts for the prediction phase.
At test time, it uses this procedure to adapt itself to new tasks from $p(\tau)$.

We define and investigate two forms of conditioning information in this work (\S\ref{sec:info}).
The first uses the gradient of the model's prediction loss with respect to network pre-activations, computed on the description data; the second replaces the loss gradient with direct feedback alignment \cite{lillicrap2016random,nokland2016direct}.
The direct feedback information is computationally cheaper and leads to competitive or superior performance in our experiments.
Note that other sources of conditioning are possible.

Our proposed neuron-level adaptation has several advantages over previous methods for metalearning that adapt the connections between neurons, for instance via fast weights \cite{pmlr-v70-munkhdalai17a} or an optimizer \cite{finn2017model,Sachin2017}. First, it is more efficient computationally, since the number of neurons is generally much less than the number of weight parameters (e.g., the number of weights scales quadratically in the number of neurons per layer for a fully connected network).
%Furthermore, neural activity can be modified straightforwardly over different time-scales, whereas weights cannot. % within separate task and objective contexts.
Second, conditionally shifted neurons can be incorporated into various neural architectures, including convolutional and recurrent networks, without special modifications to suit the structure of such models.

After describing the details of our framework, we demonstrate experimentally that ResNet \cite{he2016deep} and deep LSTM \cite{hochreiter1997long} models equipped with CSNs achieve 56.88\% and 71.94\% accuracy on the standard Mini-ImageNet 1- and 5-shot benchmarks, and 41.25\%, 52.1\%, and 57.8\% accuracy on Penn Treebank 1-, 2-, and 3-shot language modeling tasks. These results mark a significant improvement over the previous state of the art.

Our primary contributions in this paper are as follows: (i) we propose a generic neural mechanism, \emph{conditionally shifted neurons}, by which learning systems can adapt on the fly; (ii) we introduce \emph{direct feedback} as a computationally inexpensive metalearning signal; and (iii) we implement and evaluate conditionally shifted neurons in several widely-used neural network architectures.\footnote{Code and data will be available at https://aka.ms/csns}

\section{Conditionally Shifted Neurons}

\begin{figure*}[ht]
% \vskip 0.2in
\begin{center}
\includegraphics[width=0.8\textwidth]{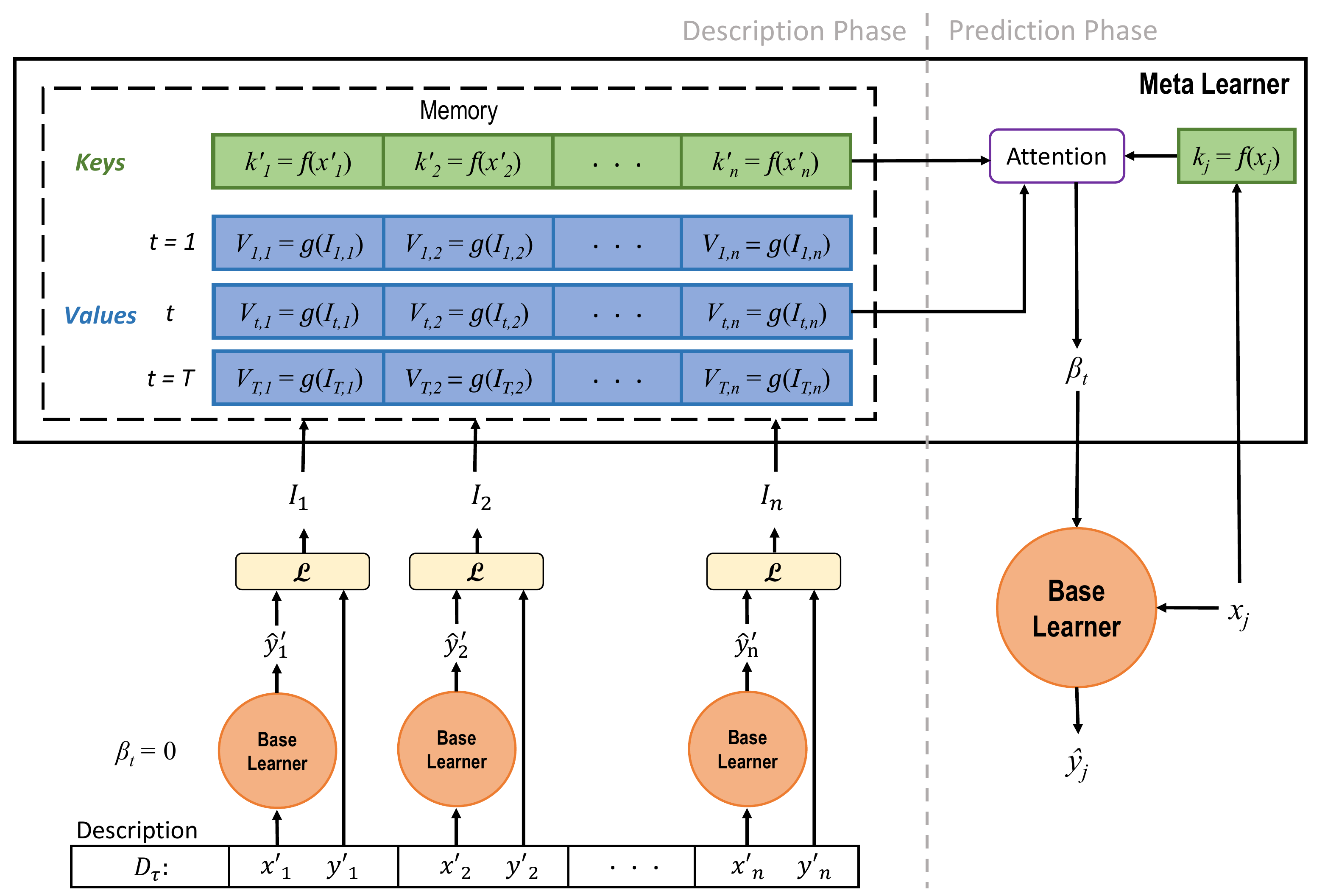}
\caption{Schematic illustration of our model with conditionally shifted neurons. In the description phase, the meta learner populates working memory with keys and values, based on the base learner's performance on the task description; in the prediction phase, the meta learner retrieves task-specific shifts from memory through key-based attention and feeds them to the base learner to adapt it to the task.
}
\label{fig:adann}
\end{center}
% \vskip -0.2in
\end{figure*}

The core idea of \textit{conditionally shifted neurons} is to modify a network's activation values on the fly, by shifting them as a function of auxiliary conditioning information.
A layer with CSNs takes the following form:
\begin{equation}
h_t = \begin{cases}
    \label{eq:adaN_t}
    \sigma (a_t) + \sigma(\beta_t) & t \neq T \\
    %\label{eq:adaN_T}
    \softmax(a_{t} + \beta_{t}) & t = T
\end{cases}
\end{equation}
for hidden layer $t$ or output layer $T$ (which represents a probability distribution). The pre-activation vector $a_t \in \mathbb{R}^{L_t}$, for a layer with $L_t$ neurons, can take various forms depending on the network architecture (fully connected, convolutional, etc.).
The nonlinear function $\sigma$ computes an element-wise activation.
$\beta_t \in \mathbb{R}^{L_t}$ is the layer-wise conditional shift vector, determined from layer-wise conditioning information $I_t$ (defined in \S\ref{sec:info}).

To implement a model with CSNs, we must define functions that extract and transform the conditioning information $I_t$ into the shifts $\beta_t$.
For this we build on the \emph{MetaNet} architecture of \citet{pmlr-v70-munkhdalai17a}.
MetaNet consists of a base learner plus a shared meta learner with working memory. For each task $\tau$, MetaNet processes the task description $D_{\tau} = \lbrace (x'_i,y'_i) \rbrace^n_{i=1}$ and stores relevant ``meta information'' in a key-value memory.
To classify unseen examples $x_j$ from the described task, the model queries its working memory with an attention mechanism to generate a set of fast weights; these modify the base learner, which in turn predicts labels $y_j$.

To begin, we describe model details for a fully connected feed-forward network (FFN) with CSNs.
The architecture is depicted in Figure \ref{fig:adann}.
As shown, the model factors into a \emph{base learner}, which makes predictions on inputs, and a \emph{meta learner}. The meta learner extracts conditioning information from the base learner and uses a key-value memory to store and retrieve activation shifts.
After walking through the model details, we define error gradient and direct feedback variants of the conditioning information $I_t$ (\S\ref{sec:info}).
In \S\ref{ada_resnet} and \S\ref{ada_lstm} we describe how CSNs can be added to ResNet and LSTM architectures, respectively.

\subsection{Feed-Forward Networks with Conditionally Shifted Neurons}

Our model operates in two phases: a description phase, wherein it processes the task description $D_{\tau} = \lbrace ( x'_i,y'_i ) \rbrace^n_{i=1}$, and a prediction phase, wherein it acts on unseen datapoints $x_j$ to predict their labels $y_j$.
In an episode of training or test, we sample a task from $p(\tau)$. The model then ingests the task description and uses what it learns therefrom, via the conditioning information, to make predictions on unseen task data.
%A procedural description of the base learner and the meta learner operating in the description phase and the prediction phase is as follows.

\subsubsection{Base Learner}
The base learner maps an input datapoint to its label prediction through layers described by equation~\ref{eq:adaN_t}, where
in the FFN case, the pre-activation vector $a_t$ is given by $a_t = W_t h_{t-1} + b_t$. Weight matrix $W_t$ and bias vector $b_t$ are learned parameters.

The base learner operates similarly in both phases.
During the \textbf{description phase}, the base learner's input is a datapoint $x'_i$ from $D_\tau$.
Its softmax output is an estimate $\hat{y}'_i$ for the label $y'_i$.
The conditional shifts $\beta_t$ in eq.~\ref{eq:adaN_t} are set to $0$ in this phase.

During the \textbf{prediction phase}, the base learner operates on inputs $x_j$. It receives conditional shifts $\beta_t$ from the meta learner and applies them layer-wise according to eq.~\ref{eq:adaN_t}.
Conditioned on these shifts, the base learner computes an estimate $\hat{y}_j$ for the label $y_j$.

\subsubsection{Meta Learner}
The meta learner's operation is more complicated and differs more significantly from phase to phase.

During the \textbf{description phase}, as the base learner processes $x'_i \in D_\tau$, the meta learner extracts layer-wise conditioning information for this example, $I_{t,i}$, according to eq.~\ref{meta_backprop} or \ref{meta_df}.
The meta learner uses the conditioning information to generate memory values.
These act as ``template'' conditional shifts for the task,
and are computed via the memory function $g$:
\begin{equation}
V_{t,i} = g(I_{t,i}),
\label{eq:g}
\end{equation}
where $V_{t,i} \in {\rm I\!R}^{L_t}$ encodes the shift template at layer $t$ for input $x'_i$. There are $n$ of these; we arrange them into matrix $V_t \in {\rm I\!R}^{n \times L_t}$ over the full task description. %Good point about n, adding this to intro

For parsimony, we desire a single memory function $g$ for all layers of the base learner, which may have different sizes $L_t$.
Therefore, we parameterize $g$ as a multi-layer perceptron (MLP) that operates independently on the vector of conditioning information for each neuron (defined in \S\ref{sec:info}). More sophisticated $L_t$-agnostic parameterizations for $g$ are possible, such as recurrent networks.

In parallel during the description phase, the meta learner constructs an embedded representation of the input that it uses to key the memory. This is the objective of the key function, $f$, which we parameterize here as an MLP with a linear output layer.
The key function generates, for each description input, a $d$-dimensional key vector $k_i' = f(x'_i)$.

At \textbf{prediction} time, the meta learner generates a memory query $k_j$ from input $x_j$ using the key function.
It uses $k_j$ to recall layer-wise shifts $\beta_t$ from memory via soft attention:
\begin{align}
\alpha &= \softmax_i(\cos(k_j, k_i')), \\
\beta_t &= \alpha^{\top}V_t.
\end{align}
Note that keys correspond to inputs, not base-learner layers.
The meta learner finally feeds the layer-wise shifts $\beta_t$ to the base learner to condition the computation of $\hat{y}_j$.

\subsection{Training and Test}
\label{sec:training}
We train and test the model in episodes. For each episode: we sample a training or test task from $p(\tau)$, process its description $D_\tau$, and then feed its unseen data forward to obtain their label predictions.
Training and test tasks are both drawn from the same distribution, but crucially, we partition the data such that the classes seen at training time do not overlap with those seen at test time.

Over a collection of training episodes, we optimize the model parameters end-to-end via stochastic gradient descent (SGD). Gradients are taken with respect to the (cross-entropy) task losses, $\mathcal{L}_\tau = \sum_j \mathcal{L}_\text{CE}(\hat{y}_j, y_j)$.
In this scheme, the model's computational graph contains the operations for processing the description, like the transformation of conditioning information and the generation of memory keys and values. Parameters of these operations are also optimized.

\subsection{Conditioning Information}
\label{sec:info}
%Now we define the two variants of conditioning information $I_t$ that we investigate in this work.
\paragraph{Error gradient information}
Inspired by the success of MetaNets, we first consider gradients of the base learner's loss on the task description as the conditioning information.
To compute these error gradients we apply the chain rule and the standard backpropagation algorithm to the base learner.
%Recall that conditioning information is extracted during the \textbf{description phase} only.
Given a true label $y'_i$ from the task description and the model's corresponding label prediction $\hat{y}'_i$, we obtain loss gradients for base-learner neurons at layer $t$ as 
\begin{equation}
\nabla_{t,i} = \frac{\partial \mathcal{L}(\hat{y}'_i, y'_i)}{\partial a_t},
\end{equation}
where $a_t$ is the $L_t$-dimensional vector of pre-activations at layer $t$, $\nabla_{t,i}$ has the same size, and we denote with $\mathcal{L}(\cdot)$ a loss function (such as the cross entropy loss on the labels). Note that $\mathcal{L}$ here is \emph{not} the target of optimization via SGD.

We obtain the conditioning information $I_{t,i,\ell}$ for each neuron (indexed by $\ell$) using the gradient preprocessing formula of~\citet{andrychowicz2016learning}:
\begin{equation}
\label{meta_backprop}
I_{t,i,\ell} =
\begin{cases}
  \left( \frac{\log(|\nabla_{t,i,\ell}|)}{p},~\text{sgn}(\nabla_{t,i,\ell})\right) & \text{if }|\nabla_{t,i,\ell}| \geq e^{-p}\\    
  (-1, e^p \nabla_{t,i,\ell}) & \text{otherwise}
\end{cases}
\end{equation}
where $\text{sgn}$ is the signum function and we set $p=7$.
We use this preprocessing to smooth variation in $I_{t,i,\ell}$, since gradients with respect to different base-learner activations can have very different magnitudes.
By eq.~\ref{meta_backprop}, each neuron obtains a $2$-dimensional vector of conditioning information. In this case, we can interpret eq.~\ref{eq:adaN_t} as a one-step, transformed gradient update on the neuron activations via $\beta_t$. ``Raw'' gradients are transformed through preprocessing, the memory read and write operations, and the nonlinearity $\sigma$.

Because backpropagation is inherently sequential, this information is expensive to compute. It becomes increasingly costly for deeper networks, such as RNNs processing long sequences.

\paragraph{Direct feedback information}
Direct feedback (DF) information is inspired by feedback alignment methods \cite{lillicrap2016random,nokland2016direct} and biologically plausible deep learning \cite{bengio2015towards}. We obtain the DF information for base-learner neurons at layer $t$ as
\begin{equation}
\label{meta_df}
I_{t,i,\ell} =  \sigma'(a_{t,\ell}) \cdot (\hat{y}'_i - y'_i),
\end{equation}
where  $\sigma'(\cdot)$ represents the derivative of the nonlinear activation function $\sigma$ and $(\hat{y}'_i - y'_i)$ is the derivative of the cross entropy loss with respect to the softmax input. Thus, the DF conditioning information for each neuron is the derivative of the loss function scaled by the derivative of the activation function.
In the DF case, each neuron obtains a $C$-dimensional vector of information, with $C$ the number of output classes.
We can compute this conditioning information for all neurons in a network simultaneously, with a single multiplication. This is more efficient than sequentially locked backpropagation-based error gradients. Furthermore, to obtain DF information, it is sufficient that only the loss and neuron activation functions are differentiable. This is more relaxed than for backpropagation methods. We demonstrate the effectiveness of both conditioning variants in \S\ref{sec_results}.

\subsection{Deep Residual Networks with CSNs}
\label{ada_resnet}
For ResNets \cite{he2016deep} we incorporate conditionally shifted neurons into the output of a residual block. Let us denote the residual block as ResBlock, which is defined as follows:
\begin{align*}
%h^1 &= \sigma(\text{BN}(\text{conv}(x)))\\
%h^2 &= \sigma(\text{BN}(\text{conv}(h^1)))\\
%
%h^1 &= \relu(\text{BN}(\text{conv}(x)))\\
%h^2 &= \relu(\text{BN}(\text{conv}(h^1)))\\
%h^3 &= \text{BN}(\text{conv}(h^2))\\
%h^4 &= \text{BN}(\text{conv}(x))\\
h^1 &= \relu(\text{conv}(x))\\
h^2 &= \relu(\text{conv}(h^1))\\
h^3 &= \text{conv}(h^2)\\
h^4 &= \text{conv}(x)\\
a_t &= h^3 + h^4
\end{align*}
where $x$ and $a_t$ are the inputs to the block and the output pre-activations, respectively. Function $\text{conv}$ denotes a convolutional layer, which may optionally be followed by 
%while BN represents 
a batch normalization \cite{ioffe2015batchnorm} layer. The activations $h_t$ of the CSNs for the ResBlock are computed as:
\begin{equation*}
%h_t = \sigma(\text{ResBlock}(x)) + \sigma(\beta_{t})
h_t = \sigma(a_t) + \sigma(\beta_{t})
\end{equation*}
where $\beta_{t}$ is the task-specific shift retrieved from the memory, constructed based on the activation values $a_t$ analogously to the FFN case; i.e., the conditioning information is computed for neurons at the output of each residual block. We stack several residual blocks with CSNs to construct a deep adaptive ResNet model. 
We use the $\relu$ function as the nonlinearity $\sigma$ in this model.

\subsection{ Long Short-Term Memory Networks with CSNs}
\label{ada_lstm}
Given the current input $x_t$, the previous hidden state $h_{t-1}$, and the previous memory cell state $c_{t-1}$, an LSTM model with CSNs computes its gates, new memory cell states, and hidden states at time step $t$ with the following update rules:
\begin{align*}
i_t &= \text{Sigmoid}(W_{i} [x_t;h_{t-1}] + b_{i})\\
f_t &= \text{Sigmoid}(W_{f}[x_t;h_{t-1}] + b_{f})\\
o_t &= \text{Sigmoid}(W_{o} [x_t;h_{t-1}] + b_{o})\\
%c_t &= \tanh (W_{v} [x_t;h_{t-1}] + b_{v}) \odot i_t + c_{t-1} \odot f_t\\
%h_t &= (\tanh(c_t) + \tanh(\beta_{t})) \odot o_t
c_t &= \sigma (W_{v} [x_t;h_{t-1}] + b_{v}) \odot i_t + c_{t-1} \odot f_t\\
h_t &= (\sigma(c_t) + \sigma(\beta_{t})) \odot o_t
\end{align*}
where $\odot$ represents element-wise multiplication, $[.;.]$ is concatenation, and $\beta_{t}$ is the task-specific shift from the memory. In the LSTM case, the memory is constructed by processing conditioning information extracted from the memory cell $c_t$. By stacking such layers together we build a deep LSTM model that adapts across both depth and time. 
We use the $\tanh$ function as the nonlinearity $\sigma$ in this model.

\section{Related Work}
\label{related work}

Among the many problems in supervised, reinforcement, and unsupervised learning that can be framed as metalearning, few-shot learning has emerged as a natural and popular test bed.
Few-shot supervised learning refers to a scenario where a learner is introduced to a sequence of tasks, where each task entails multi-class classification given a single or very few labeled examples per class. A key challenge in this setting is that the classes or concepts vary across the tasks; thus, models require a capacity for rapid adaptation in order to recognize new concepts on the fly.

Few-shot learning problems were previously addressed using metric learning methods \cite{koch2015siamese}. Recently, there has been a shift towards building flexible models for these problems within the learning-to-learn paradigm \cite{mishra2017meta,santoro2016meta}. \citet{vinyals2016matching} unified the training and testing of a one-shot learner under the same procedure and developed an end-to-end, differentiable nearest-neighbor method for one-shot learning. More recently, one-shot optimizers were proposed by \citet{Sachin2017,finn2017model}. The MAML framework~\citep{finn2017model} learns a parameter initialization from which a model can be adapted rapidly to a given task using only a few steps of gradient updates. To learn this initialization it makes use of more sophisticated second-order gradient information. Here we harness only first-order gradient information, or the simpler direct feedback information.

As highlighted, the architecture of our model with conditionally shifted neurons is closely related to Meta Networks \cite{pmlr-v70-munkhdalai17a}. The MetaNet modifies synaptic connections (weights) between neurons using fast weights \cite{schmidhuber1987,hinton1987using} to implement rapid adaptation. While MetaNet's fast weights enable flexibility, it is very expensive to modify these weights when the connections are dense. Neuron-level adaptation as proposed in this work is significantly more efficient while lending itself to a range of network architectures, including ResNet and LSTM. Other previous work on metalearning has also formulated the problem as two-level learning: specifically, ``slow'' learning of a meta model across several tasks, and ``fast'' learning of a base model that acts within each task \cite{schmidhuber1987,bengio1990learning,hochreiter2001learning,mitchell1993explanation,vilalta2002perspective,mishra2017meta}.
\citet{schmidhuber1993self} discussed the use of network weight matrices themselves for continuous adaptation in dynamic environments.

Viewed as a form of feature-wise transformation, CSNs are closely related to conditional normalization techniques \cite{lei2015predicting,dumoulin2016learned,ghiasi2017exploring,de2017modulating,perez2017film}. FiLM~\cite{perez2017film}, the most similar such approach which was inspired by \citet{dumoulin2016learned} and \citet{ghiasi2017exploring}, modulates CNN \emph{feature maps} using global scale and shift operations conditioned on an auxiliary input modality. %\footnote{The pre-activation $a_{t-1}$ is shifted as $\gamma a_{t-1} + \beta$, where $\gamma$ and $\beta$ are learned functions of the auxiliary input.}
In contrast, CSNs apply shifts to individual neurons' activations, locally, and this modification is based on the model's behavior on the task description rather than the input itself.

In the case of gradient-based conditioning information, our approach can be viewed as a synthesis of a conditional normalization model (in the style of FiLM) with a learned optimizer (in the style of \citet{andrychowicz2016learning}).
Specifically, the learned memory and key functions, $g$ and $f$, transform error gradients into the conditioning shifts $\beta_t$, which are then applied like a one-step update to the activation values.
A CSN model uses this learned optimizer on the fly.

\section{Experimental Evaluation}
\label{sec_results}
\begin{table*}[t] %%%%%%%%%%%%%%% OMNIGLOT TABLE %%%%%%%%%%%%%%% 
%  \caption{Test accuracy on Omniglot few-shot classification. $\nabla$: error gradient-based conditioning information, DF: direct feedback conditioning information.}
  \caption{Omniglot few-shot classification test accuracy for error gradient ($\nabla$) and direct feedback (DF) conditioning information.}
  \label{tab:omni}
  \small
  \centering
  \begin{tabular}{lcccc}
    \toprule
    {} & \multicolumn{2}{c}{\bf 5-way} & \multicolumn{2}{c}{\bf 20-way}   \\
    \cmidrule(l{3pt}r{3pt}){2-3} \cmidrule(l{3pt}r{3pt}){4-5}
    \bf Model & \bf 1-shot & \bf 5-shot & \bf 1-shot & \bf 5-shot             \\
    \midrule
    Siamese Net \cite{koch2015siamese} & 97.3 & 98.4 & 88.2 & 97.0 \\
    MANN \cite{santoro2016meta} & 82.8 & 94.9 & - & - \\
    Matching Nets \cite{vinyals2016matching} & 98.1 & 98.9 & 93.8 & 98.5 \\
    MAML \cite{finn2017model} & \bf 98.7 $\pm$ 0.4 & \bf 99.9 $\pm$ 0.3 & 95.8 $\pm$ 0.3 & 98.9 $\pm$ 0.2 \\
    MetaNet \cite{pmlr-v70-munkhdalai17a} & \bf 98.95 & - & 97.0 & - \\
    TCML \cite{mishra2017meta} & \bf 98.96 $\pm$ 0.2 & 99.75 $\pm$ 0.11 & \bf 97.64 $\pm$ 0.3 & \bf 99.36 $\pm$ 0.18 \\
    \midrule
%     adaCNN ($\nabla$) & \bf 98.45 $\pm$ 0.32 & 99.44 $\pm$ 0.24 & 96.51 $\pm$ 0.45 & 98.89 $\pm$ 0.21 \\
%     adaCNN (DF) & 98.32 $\pm$ 0.33 & 99.25 $\pm$ 0.26 & 96.24 $\pm$ 0.6 & 98.13 $\pm$ 0.37\\
     adaCNN ($\nabla$) & 98.41 $\pm$ 0.16 & 99.27 $\pm$ 0.12 & 95.95 $\pm$ 0.43 & 98.48 $\pm$ 0.06 \\
%     adaCNN (DF) & 98.42 $\pm$ 0.21 & 99.37 $\pm$ 0.28 & 96.12 $\pm$ 0.31 & 98.43 $\pm$ 0.052 \\
     adaCNN (DF) & 98.42 $\pm$ 0.21 & 99.37 $\pm$ 0.28 & 96.12 $\pm$ 0.31 & 98.43 $\pm$ 0.05 \\
    \bottomrule
  \end{tabular}
\end{table*}
\begin{table*}[t] %%%%%%%%%%%%%%% MINI-IMAGENET TABLE %%%%%%%%%%%%%%%
%  \caption{Test accuracy on Mini-ImageNet few-shot classification. $\nabla$: gradient-based conditioning information, DF: direct feedback conditioning information.}
\caption{Mini-ImageNet few-shot classification test accuracy for error gradient ($\nabla$) and direct feedback (DF) conditioning information.}
  \label{tab:mini}
  \small
  \centering
  \begin{tabular}{lcc}
    \toprule
    {} & \multicolumn{2}{c}{\bf 5-way} \\
    \cmidrule(l{3pt}r{3pt}){2-3} 
    \bf Model & \bf 1-shot & \bf 5-shot \\
    \midrule
    Matching Nets \cite{vinyals2016matching} & 43.6 & 55.3 \\
    MetaLearner LSTM \cite{Sachin2017} & 43.4 $\pm$ 0.77 & 60.2 $\pm$ 0.71 \\
    MAML \cite{finn2017model} & 48.7 $\pm$ 1.84 & 63.1 $\pm$ 0.92 \\
    MetaNet \cite{pmlr-v70-munkhdalai17a} &  49.21 $\pm$ 0.96 & - \\
    \midrule
    adaCNN ($\nabla$) & 48.26 $\pm$ 0.63 & 62.80 $\pm$ 0.41 \\
    adaCNN (DF) & 48.34 $\pm$ 0.68 & 62.00 $\pm$ 0.55 \\
    \midrule \midrule
    TCML \cite{mishra2017meta} & 55.71 $\pm$ 0.99 & 68.88 $\pm$ 0.92 \\
    \midrule
%    adaResNet ($\nabla$) & 56.616 $\pm$ 0.69 & 71.69 $\pm$ 0.67 \\
    adaResNet ($\nabla$) & 56.62 $\pm$ 0.69 & 71.69 $\pm$ 0.67 \\
    adaResNet (DF) & \bf 56.88 $\pm$ 0.62 & \bf 71.94 $\pm$ 0.57 \\
    \bottomrule
  \end{tabular}
\end{table*}
\begin{table*}[t] %%%%%%%%%%%%%%% PTB TABLE %%%%%%%%%%%%%%%
%  \caption{One-shot language modeling test accuracy. $\nabla$: gradient-based conditioning information, DF: direct feedback conditioning information}
\caption{Penn Treebank few-shot classification test accuracy for error gradient ($\nabla$) and direct feedback (DF) conditioning information.}
  \label{tab:lm}
  \small
  \centering
  \begin{tabular}{lccc}
    \toprule
    {} & \multicolumn{3}{c}{\bf 5-way (400 random/all-inclusive)} \\
    \cmidrule(l{3pt}r{3pt}){2-4} 
    \bf Model & \bf 1-shot & \bf 2-shot & \bf 3-shot \\
    \midrule
    LSTM-LM oracle \cite{vinyals2016matching} & 72.8 & 72.8 & 72.8 \\
    Matching Nets \cite{vinyals2016matching} & 32.4 & 36.1 & 38.2 \\
    \midrule
%     2-layer LSTM-LM baseline & \multicolumn{3}{c}{59.8/61.5} \\
     2-layer LSTM + adaFFN ($\nabla$) & 32.55/33.2 &	44.15/46.0 &	50.4/51.7 \\
     1-layer adaLSTM ($\nabla$) & 36.55/37.7 &	43.25/44.6 &	50.7/52.1 \\
     2-layer adaLSTM ($\nabla$) & \bf 43.1/43.0 & \bf	52.05/54.2 & \bf	57.35/58.4 \\
     2-layer LSTM + adaFFN (DF) & 33.65/35.3 &	46.6/47.8 &	51.4/52.6 \\
    1-layer adaLSTM (DF) & 36.35/36.3 &	41.6/43.4 &	49.1/50.1 \\
    2-layer adaLSTM (DF) & \bf 41.25/43.2 & \bf	52.1/52.9 & \bf	57.8/58.8 \\
    \bottomrule
  \end{tabular}
\end{table*}
We evaluate the proposed CSNs on tasks from the vision and language domains. Below we describe the datasets we evaluate on and the according preprocessing steps, followed by test results and an ablation study.

\subsection{Few-shot Image Classification}
In the vision domain, we used two widely adopted few-shot classification benchmarks: the Omniglot and Mini-ImageNet datasets.

{\bf Omniglot} consists of images from 1623 classes from 50 different alphabets, with only 20 images per class \cite{lake2015human}. As in previous studies, we randomly selected 1200 classes for training and 423 for testing and augmented the training set with 90, 180 and 270 degree rotations. We resized the images to $28 \times 28$ pixels for computational efficiency.

For the Omniglot benchmark we performed 5- and 20-way classification tests, each with one or five labeled examples from each class as the description $D_\tau$. We use a convolutional network (CNN) with 64 filters as the base learner. This network has 5 convolutional layers, each of which uses $3 \times 3$ convolutions followed by the ReLU nonlinearity and a $2 \times 2$ max-pooling layer. Convolutional layers are followed by a fully connected (FC) layer with softmax output. Another CNN with the same architecture is used for the key function $f$. We use CSNs in the last four layers of the CNN components, referring to this model as ``adaCNN.'' Full implementation details can be found in Appendix~\ref{sec:imp_details}.

Table \ref{tab:omni} shows that our adaCNN model achieves competitive, though not state-of-the-art, results on the Omniglot tasks. There is an obvious ceiling effect among the best performing models as accuracy saturates near 100\%.

{\bf Mini-ImageNet} features $84 \times 84$-pixel color images from 100 classes (64/16/20 for training/validation/test splits) and each class has 600 exemplar images. We ran our experiments on the class subset released by \citet{Sachin2017}.
Compared to Omniglot, Mini-ImageNet has fewer classes (100 vs 1623) with more labeled examples provided of each class (600 vs 20). Given this larger number of examples,  we evaluated a similar adaCNN model with 32 filters as well as a model with more sophisticated ResNet components (``adaResNet'') on the Mini-ImageNet 5-way classification tasks. The ResNet architecture follows that of TCML~\citep{mishra2017meta} with two exceptions due to memory constraints. Instead of two $1 \times 1$ convolutional layers with 2048 and 512 filters we use only a single such layer with 1024 filters, and the ReLU nonlinearity instead of its leaky variant. We incorporate CSNs into the last two residual blocks as well as the two fully connected output layers. Full implementation details can be found in Appendix~\ref{sec:imp_details}.

For every 400 training tasks, we tested the model for another 400 tasks sampled from the validation set. If the model performance exceeded the previous best validation result, we applied it to the test set. Following previous approaches that we compare with in Table \ref{tab:mini}, we sampled another 400 tasks randomly from the test set to report model accuracy.

Unlike Omniglot, there remains significant room for improvement on Mini-ImageNet.
As shown in Table \ref{tab:mini}, on this more challenging task, CNN-based models with conditionally shifted neurons achieve performance just below that of the best CNN-based approaches like MAML
and MetaNet
(recall that these modify weight parameters rather than activation values). The more sophisticated adaResNet model, on the other hand, achieves state-of-the-art results. The best-performing adaResNet (DF) yields almost 10\% improvement over the corresponding adaCNN model and improves over the previous best result of TCML by 1.16\% and 3.06\% on the one and five shot 5-way classification tasks, respectively. Note that TCML likewise uses a ResNet architecture.
%Somewhat surprisingly, direct feedback conditioning information outperformed the gradient-based variant on the Mini-ImageNet task.
The best accuracy among five different seed runs of the adaResNet with DF conditioning was 72.91\% on the five-shot task.

\subsection{Few-shot Language Modeling}

To evaluate the effectiveness of recurrent models with conditionally shifted neurons, we ran experiments on the few-shot \textbf{Penn Treebank} (PTB) language modeling task introduced by \citet{vinyals2016matching}.

In this task, a model is given a query sentence with one missing word and a support set (i.e., description) of one-hot-labeled sentences that also have one missing word each. One of the missing words in the description set is identical to that missing from the query sentence. The model must select the label of this corresponding sentence.

Following \citet{vinyals2016matching}, we split the PTB sentences into training and test such that, for the test set, target words for prediction and the sentences in which they appear are unseen during training. Concretely, we removed the test target words as well as sentences containing those words from the training data. This process necessarily reduces the training data and increases out-of-vocabulary (OOV) test words. We used the same 1000 target words for testing as provided by \citet{vinyals2016matching}.

We evaluated two models with conditionally shifted neurons on 1-, 2-, and 3-shot language modelling (LM) tasks. In both cases, we represent words with randomly initialized dense embeddings. For the first model we stacked a 3-layer feed-forward net with CSNs (adaFFN) on top of an LSTM network (LSTM+adaFFN) at each prediction timestep. In this model, only the adaFFN can adapt to the task while it processes the hidden state of the underlying LSTM. The LSTM encoder builds up the context for each word and provides a generic (non-task-specific) representation to the adaFFN. Both components are trained jointly.

The second model we propose for this task is more flexible, an LSTM with conditionally shifted neurons in the recurrence (adaLSTM). This entire model is adapted with task-specific shifts at every time step. For few-shot classification output, a softmax layer with CSNs is stacked on top of the adaLSTM. Comparing LSTM+adaFFN and adaLSTM, the former is much faster since we only adapt the activations of the three feedforward layers, but it lacks full flexibility since the LSTM is unaware of the current task information. We also evaluated deep (2-layer) versions of both LSTM+adaFFN and adaLSTM models. Full implementation details can be found in Appendix~\ref{sec:imp_details}.

We used two different methods to form test tasks for evaluation. First, we randomly sampled 400 tasks from the test data and report the average accuracy. Second, we make sure to include all test words in the task formulation. We randomly partition the 1000 target words into 200 groups and solve each group as a task. In the random approach there is a chance that a word could be missed or included multiple times in different tasks. However, the random approach also enables formulation of an exponential number of test tasks.

Table \ref{tab:lm} summarizes our results. The approximate upper bound achieved by the oracle LSTM-LM of \citet{vinyals2016matching} is 72.8\%. Our best accuracy -- around 58\% on the 3-shot task -- comes using a 2-layer adaLSTM, and improves over the Matching Nets results by 11.1\%, 16.0\% and 19.6\% for 1-, 2-, and 3-shot tasks, respectively. Comparing model variants, adaLSTM consistently outperforms the standard LSTM augmented with a conditionally shifted output FFN, and deeper models yield higher accuracy. Providing more sentences for the target word increases performance, as expected.
These results indicate that our model's few-shot language modelling capabilities far exceed those of Matching Networks \cite{vinyals2016matching}. Some of this improvement surely arises from adaLSTM's recurrent structure, which is known to apply well to sequence-based tasks and in the language domain. However, it is one of the strengths of conditionally shifted neurons that they can be ported easily to various neural architectures.

Comparing direct feedback information to the gradient-based variant across the full suite of experiments, we observe overall that DF information performs competitively well. Even more positively, DF information speeds up the the model runtime considerably. For example, the 2-layer adaLSTM processed 400 test episodes in 200 seconds using gradient information vs. 160 seconds for DF information, representing a speedup of about 25\%. In Appendix \ref{sec:speed}, we compare runtimes for the adaCNN variants and MetaNet model on the Mini-ImageNet task. Even for the shallow adaCNN model, we observe 2-3 ms/task speed-up with the DF conditioning information.

\subsection{Ablation Study}

\begin{figure}[h] %%%%%%%%%%%%%%% MINI-IMAGENET ABLATION FIGURE %%%%%%%%%%%%%%%
% \vskip 0.2in
\begin{center}
\centerline{\includegraphics[width=0.4\textwidth, trim={2cm 2.5cm 2.2cm 2cm},clip]{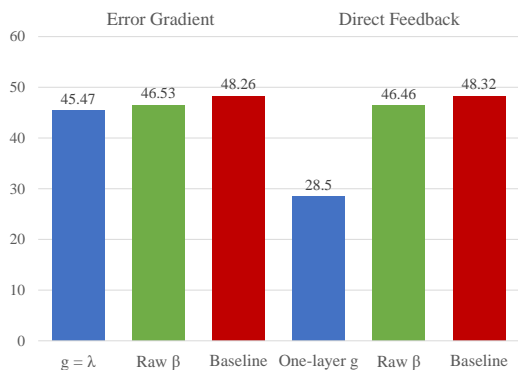}}
\caption{Model ablation for adaCNN tested on the Mini-ImageNet one-shot task. Blue: $g$ as a scalar multiplier of $\nabla_{t,i}$ (gradient case) or perceptron (DF case); Green: $\beta$ without normalization; Red: baseline model.}
\label{fig:abl_mini}
\end{center}
 \vskip -.2in
% \vskip -1.0in
\end{figure}

\begin{figure}[h] %%%%%%%%%%%%%%% LM ABLATION FIGURE %%%%%%%%%%%%%%%
% \vskip 0.2in
\begin{center}
\centerline{\includegraphics[width=0.4\textwidth, trim={2cm 2.5cm 2.2cm 2cm},clip]{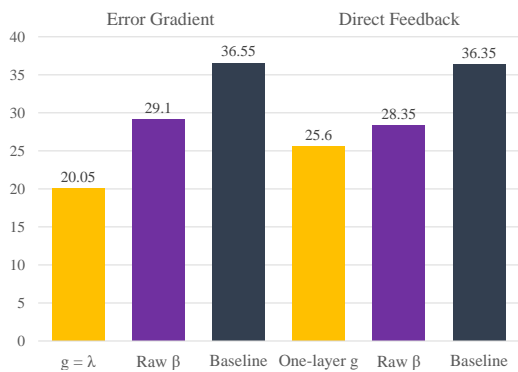}}
\caption{Model ablation on the one-shot language modeling task, for a single layer adaLSTM. We report average accuracy on 400 random test tasks. Yellow: $g$ as a scalar multiplier of $\nabla_{t,i}$ (gradient case) or perceptron (DF case); Violet: $\beta$ without normalization; Grey: baseline model.}
\label{fig:abl_lm}
\end{center}
 \vskip -.2in
% \vskip -1.0in
\end{figure}

To better understand our model, we performed an ablation study on adaCNN trained on Mini-ImageNet and the 1-layer adaLSTM trained on PTB.
Results are shown in Figure \ref{fig:abl_mini} and Figure \ref{fig:abl_lm}, respectively.

% begin{figure}[ht]
% begin{figure}[ht]

Our first ablation was to determine the effect of normalizing the task shifts $\beta_t$ through the nonlinear activation function $\sigma$. Ablating the activation function from eq.~\ref{eq:adaN_t} and simply adding $\beta_t$ resulted in a slight performance drop on the Mini-ImageNet task and a significant decrease (around 7\%) on the one-shot LM task, for both variants of conditioning information.
We similarly tried adding $\beta_t$ directly to the pre-activations $a_t$ inside the nonlinearity $\sigma$. On Mini-ImageNet, this variant of adaCNN achieved 47.8\% and 48.09\% accuracy with gradient and DF conditioning information, respectively, which is competitive with the baseline.
However, adaLSTM performance decreased more significantly to 30.7/31.8\% and 33.25/34.05\% on the LM task.
We conclude that squashing the conditional shift $\beta_t$ to the same range as the neuron's standard activation value is beneficial.

Our second ablation evaluates variations on the function $g$ for transforming the conditioning information $I_{t,i}$ into the memory values $V_{t,i}$.
In the case of gradient-based conditioning information, we remove the preprocessing of eqn.~\ref{meta_backprop} and replace the learned MLP with a simple learned scalar, $\lambda \in \mathbb{R}$, that multiplies the gradient vector $\nabla_{t,i}$.
In this case we can more clearly interpret the conditional shift as a one-step gradient update on the activation values (although this update is still modulated by the memory read procedure and the function $\sigma$). As shown in Figure \ref{fig:abl_mini}, the adaCNN model with learned scaling loses about 3 percentage points of accuracy on the image classification task.
However, as per Figure \ref{fig:abl_lm}, adaLSTM performance plummets with learned scaling, dropping to 20\% test accuracy (random chance).

For direct feedback conditioning information, we cannot use a scalar parameter because we have a $C$-dimensional information vector for each neuron (recall \S\ref{sec:info}). We therefore parameterize $g$ as a one-layer perceptron in this case rather than a deep MLP.
%\tred{Tsendee, please confirm there's no activation function in this 1-layer perceptron, if so, we should mention it here.}
Using a simple perceptron to process the direct feedback information decreased test accuracy significantly on the Mini-ImageNet and LM tasks (drops of over 10\%). This highlights that a deeper mapping function is crucial for processing DF conditioning information.

%We finally attempted to use a fixed word-embedding layer for the LM task. A 2-layer adaLSTM ($\nabla$) model with fixed embeddings performed quite well, obtaining 42.65/43.7\% and 51.45/51.7\% accuracy on 1- and 2-shot problems; this is competitive with the results given in Table \ref{tab:lm}.

\section{Conclusion}
We introduced \emph{conditionally shifted neurons}, a mechanism for rapid adaptation in neural networks. 
Conditionally shifted neurons are generic and easily incorporated into various neural architectures. They are also computationally efficient compared to alternative metalearning methods that adapt synaptic connections between neurons.
We proposed two variants of conditioning information for use with CSNs, one based on error gradients and another based on feedback alignment methods. The latter is more efficient because it does not require a sequential backpropagation procedure, and achieves competitive performance with the former.
We demonstrated empirically that models with conditionally shifted neurons improve the state of the art on metalearning benchmarks from the vision and language domains.
%\tred{Future work??}

%\clearpage

\bibliography{example_paper}
\bibliographystyle{icml2018}

\clearpage

\appendix

\section{Additional Implementation Details}
\label{sec:imp_details}

The hyperparameters for our models are listed in Tables \ref{tab:hyper-img} and \ref{tab:hyper-lm}. Key size $d$ was 64 throughout all experiments. A dropout rate of 0.2 was applied to each layer of adaFFN. For the other adaptive models, the input dropout rate was set to 0.2 or 0.0. The dropout for the last two layers were varied as shown in Table \ref{tab:hyper-img} and \ref{tab:hyper-lm}. Due to memory constraints, we used adaLSTM with a smaller number of hidden units (i.e., 200 vs 300) for deep models when applying to 3-shot tasks.

The neural network weights were initialized using \citet{he2015delving}'s method. We set the hard gradient clipping threshold for adaCNN model to 10. % although we did not attempt to tune this value.
No gradient clipping was performed for the other models. We listed the setup for optimizers in Table \ref{tab:hyper-img} and \ref{tab:hyper-lm}. For Adam optimizer, the rest of the hyperparameters were set to their default values (i.e., $\beta_1 = 0.9$, $\beta_2 = 0.999$, and $\epsilon = 10^{-8}$).

Although different parameterizations for the meta learner function $g$ may improve the performance, for simplicity we used a 3-layer MLP with $\relu$ activation with 20 or 40 units per layer. This MLP acts coordinate-wise and processes conditioning information for each neuron independently.

Empirically, we found that selecting the vector from $V_t$ corresponding to the key $k'_i$ with maximum cosine similarity to the query $k_j$ (hard attention) gave similar performance to soft attention.

We occasionally observed difficulty in optimizing the LSTM+adaFFN models, often seeing no improvement in the training loss from certain initializations. Decreasing the learning rate and in case of DF information applying dropouts to adaFFN layers helped training this model.

Models were implemented using the Chainer~\citep{chainer_learningsys2015} framework\footnote{https://chainer.org/}.

\begin{table}[h]
  \centering
  \caption{Hyperparameters for few-shot image classification tasks}
  \label{tab:hyper-img}
  \small
% \centering
  \begin{tabular}{lccccc}
    \toprule
        \bf Model & \bf Layers & \bf Filters & \bf Dropout rate & \bf Optimizer \\
    \midrule
        adaCNN ($\nabla$) & 5 & 32/64 & 0.0, 0.3, 0.3 &	Adam ($\alpha$=0.001) \\
        adaCNN (DF) & 5 &	32/64 &	0.2, 0.3, 0.3 &	Adam ($\alpha$=0.001) \\
        adaResNet ($\nabla$) & 4 &	64, 96, 128, 256 &	0.2, 0.5, 0.5 & SGD with momentum (lr=0.01, m=0.9) \\
        adaResNet (DF) & 4 & 64, 96, 128, 256 &	0.2, 0.5, 0.5 & SGD with momentum (lr=0.01, m=0.9) \\
    \bottomrule
  \end{tabular}
\end{table}

\begin{table}[h]
\vskip -0.5cm
  \caption{Hyperparameters for few-shot language modelling tasks}
  \label{tab:hyper-lm}
  \small
  \centering
  \begin{tabular}{lcccc}
    \toprule
        \bf Model & \bf Hidden unit size & \bf Dropout rate & \bf Optimizer \\
    \midrule
        % 2-layer LSTM-LM baseline & 300	& [0.3, 0.3] & Adam ($\alpha$=0.001, $\beta_1$=0.9)  &			\\
        2-layer LSTM + adaFFN ($\nabla$) &	300 &	- &	Adam ($\alpha$=0.0003) \\
        2-layer LSTM + adaFFN (DF) &	300 &	0.2, 0.2, 0.2 & Adam ($\alpha$=0.0003) \\
        1-layer adaLSTM ($\nabla$) &	300 &	- &	Adam ($\alpha$=0.001) \\
        1-layer adaLSTM (DF) &	300 &	- &	Adam ($\alpha$=0.001) \\
        2-layer adaLSTM ($\nabla$) &	300, 200 &	- &	Adam ($\alpha$=0.001) \\
        2-layer adaLSTM (DF) &	300, 200 &	- &	Adam ($\alpha$=0.001) \\
    \bottomrule
  \end{tabular}
\end{table}

\clearpage

\section{Running Time Comparison with MetaNet}
\label{sec:speed}
\begin{figure}[ht] %%%%%%%%%%%%%%% TIMING FIGURE %%%%%%%%%%%%%%%
% \vskip -1.0in
% \begin{center}

\centerline{\
\includegraphics[width=0.40\textwidth, trim={2cm 2.5cm 2.2cm 2cm},clip]{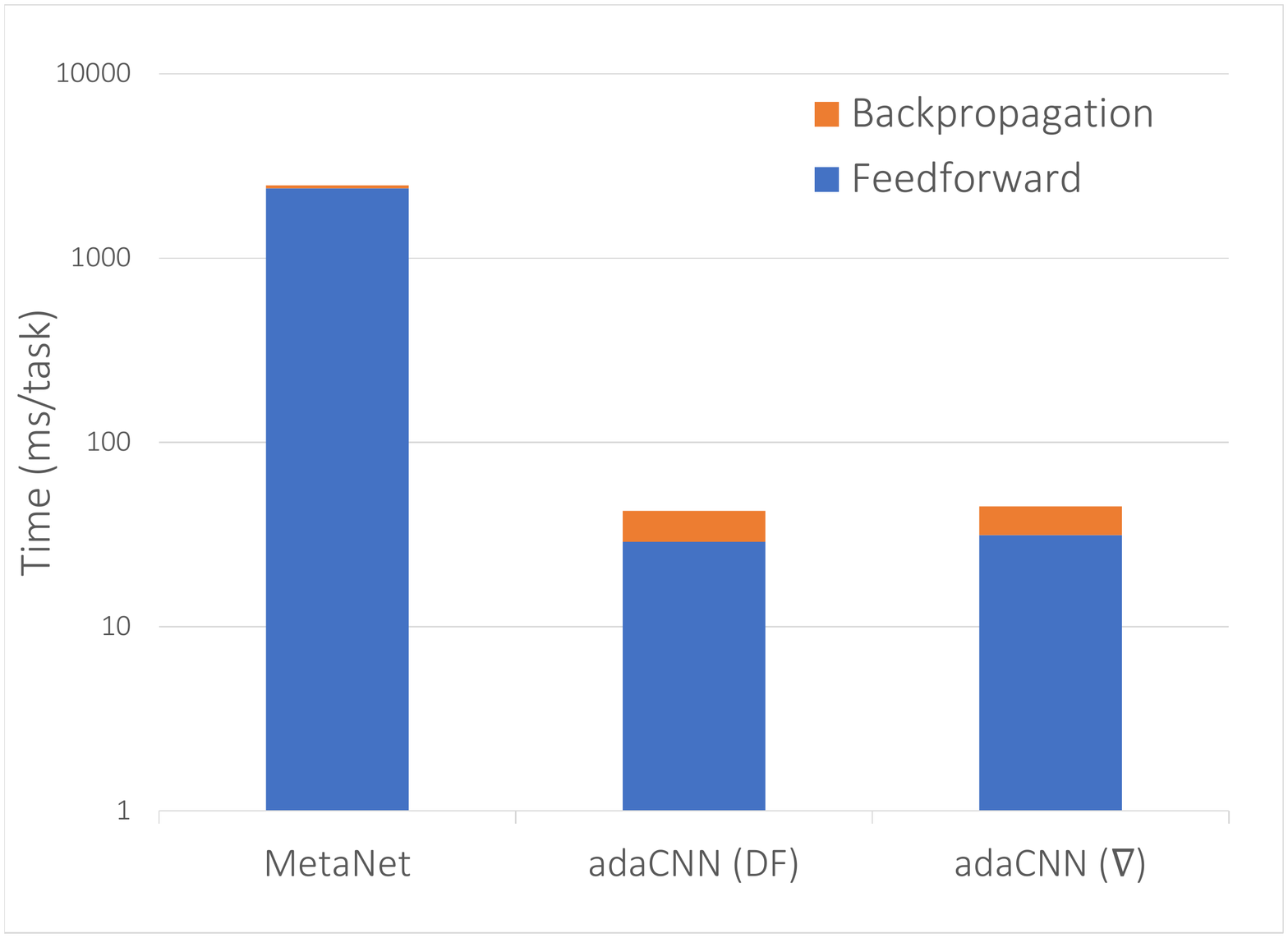}
}
\centerline{\
\includegraphics[width=0.40\textwidth, trim={2cm 2.5cm 2.2cm 2cm},clip]{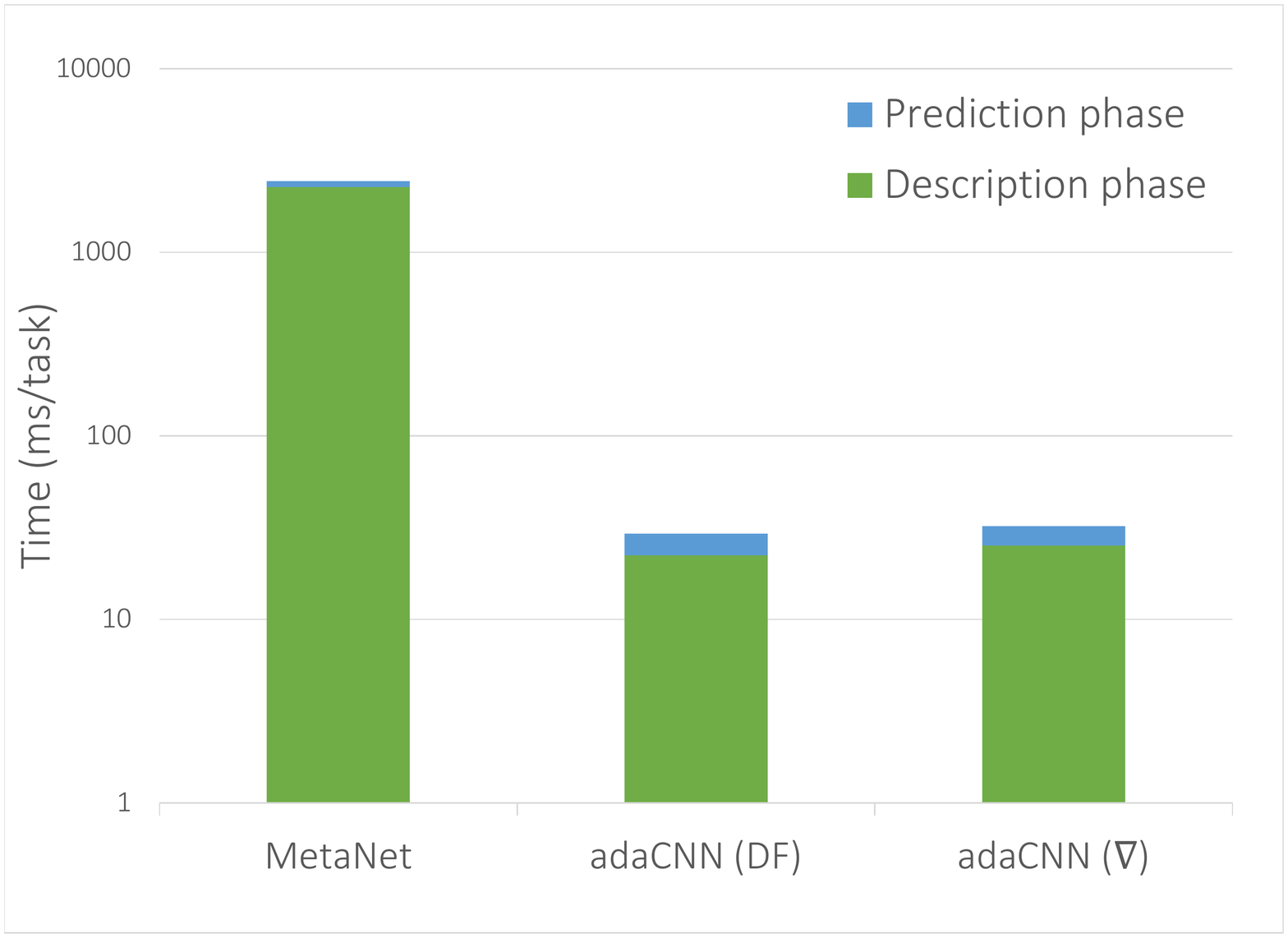}
}
\caption{Training (top) and inference (bottom) speeds of MetaNet, adaCNN variants are compared on the 1-shot, 5-way Mini-ImageNet task. 
The y-axis shows wall-clock time (ms/task) in log scale. Training time includes the feedforward computations and the parameter updates. Inference time includes computations for the description and prediction phases.}
\label{fig:timing_mini}
% \end{center}
 \vskip -.2in
% \vskip -1.0in
\end{figure}

We compared the speed of our adaCNN model variants with MetaNet model on Mini-ImageNet task. We implemented all models in the Chainer framework \cite{chainer_learningsys2015} and tested on an Nvidia Titan X GPU. In Figure~\ref{fig:timing_mini} we see that adaCNN variants are significantly faster than MetaNet while being  conceptually simpler and easier to implement.

\end{document}